\documentclass[letterpaper, 10 pt, conference]{ieeeconf}  

\IEEEoverridecommandlockouts                              

\usepackage{tikz}
\usepackage{graphicx}
\usepackage{graphics} 
\usepackage{epsfig} 
\usepackage{bbm}
\usepackage{times} 
\usepackage{amsmath} 
\usepackage{amssymb}  
\usepackage{algorithm}
\usepackage{algpseudocode}
\usepackage{graphicx, wrapfig}
\usepackage{caption}
\usepackage{subcaption}
\usepackage{bm}
\usepackage{pgfplots}
\usepackage{pgfkeys}
\usepackage[export]{adjustbox}
\usepackage{color}
\usepackage{cite}
\usepackage{multirow}
\usepackage{hyperref}
\usepackage{textcomp}
\usepackage{siunitx}
\usepackage{color}
\usepackage{dblfloatfix}
\usepackage[colorinlistoftodos]{todonotes}

\title{\LARGE \bf  Uncertainty-Aware Data Aggregation for Deep Imitation Learning}

\author{Yuchen Cui$^{1,2}$  \,  David Isele$^{2}$ \,  Scott Niekum$^{1}$ \,  Kikuo Fujimura$^{2}$
\thanks{$^{1}$Yuchen Cui and Scott Niekum are with the University of Texas at Austin, Austin, TX 78712, USA
       {\tt\small yuchencui@utexas.edu, sniekum@cs.utexas.edu}}%
\thanks{$^{2}$David Isele and Kikuo Fujimura are with Honda Research Institute USA, 375 Ravendale Dr, Mountain View, CA 94043, USA
        {\tt\small \{disele, kfujimura\}@honda-ri.com}; Yuchen Cui conducted the research during an internship at HRI. }%
}

\begin{document}

\maketitle
\thispagestyle{empty}
\pagestyle{empty}

\begin{abstract}

Estimating statistical uncertainties allows autonomous agents to communicate their confidence during task execution and is important for applications in safety-critical domains such as autonomous driving. 
In this work, we present the uncertainty-aware imitation learning (UAIL) algorithm for improving end-to-end control systems via data aggregation. UAIL applies Monte Carlo Dropout to estimate uncertainty in the control output of end-to-end systems, using states where it is uncertain to selectively acquire new training data. In contrast to prior data aggregation algorithms that force human experts to visit sub-optimal states at random, UAIL can anticipate its own mistakes and switch control to the expert in order to prevent visiting a series of sub-optimal states. Our experimental results from simulated driving tasks demonstrate that our proposed uncertainty estimation method can be leveraged to reliably predict infractions. Our analysis shows that UAIL outperforms existing data aggregation algorithms on a series of benchmark tasks. 

\end{abstract}

\begin{figure*}[ht]
\centering
\includegraphics[scale=0.7]{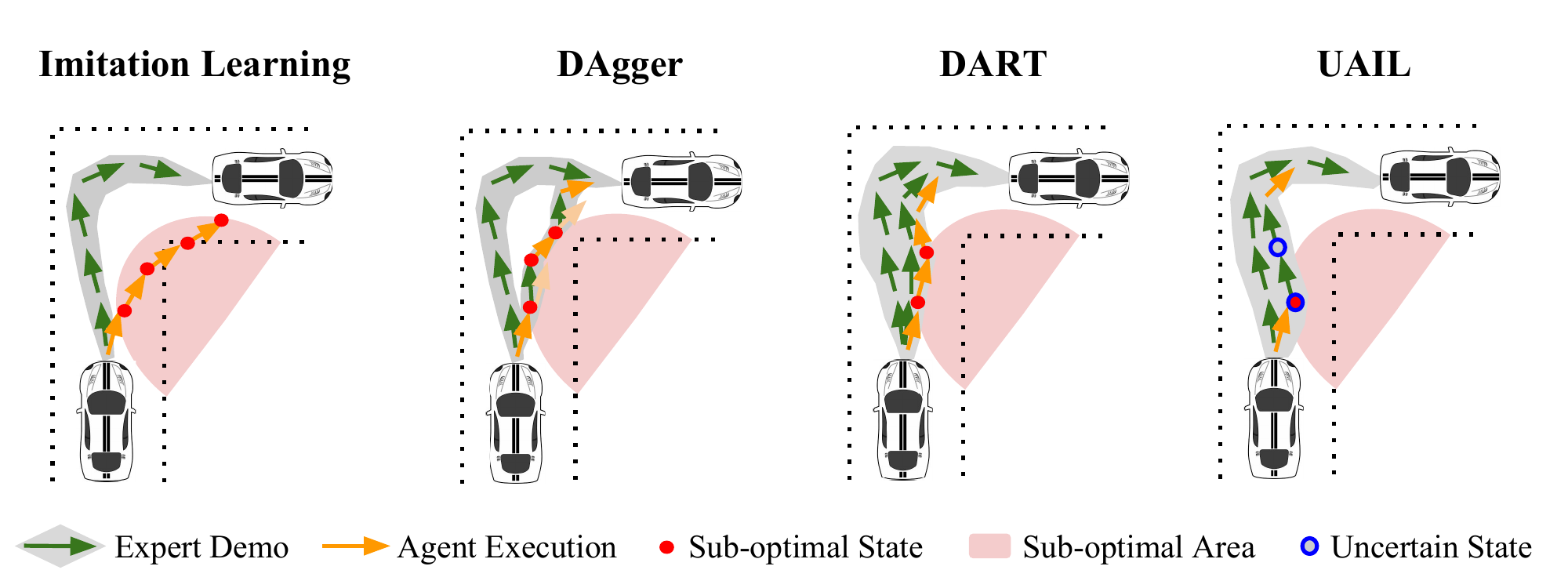}
\caption{\small Comparison of different data aggregation algorithms: pure imitation learning is off-policy and a mistake early in the trajectory propagates; DAgger is on-policy and mixes the expert's and the agent's control, forcing the expert to provide labels at suboptimal states the agent's policy visits; DART is off-policy and approximates the learned policy's error by injecting noise; UAIL is on-policy and actively switches control to the expert when the agent is uncertain.}
\label{fig:comparison}
\end{figure*}


\section{INTRODUCTION}

With recent advancement in training deep neural networks, end-to-end systems have been shown to outperform their modularized counterparts in a variety of tasks~\cite{chen2018best,maggiori2017convolutional,silver2017mastering}.  
However, end-to-end control of robotic systems remains challenging and has attracted much recent research effort~\cite{pomerleau1989alvinn,muller2006off,bojarski2017explaining,zhang2016query,levine2016end}.

One disadvantage of end-to-end learning is that it does not typically offer the same level of transparency in decision-making as simpler, more traditional systems, largely obstructing any efforts to make safety guarantees or identify failure cases in advance.

Developing methods for estimating the predictive uncertainty of end-to-end systems is one way to determine whether a learning agent is producing behaviors that should not be trusted. This work investigates how a learning agent's ability to detect uncertain states and return an \textit{``I don't know''} response can be used to predict infractions, improve the quality of the data collected, and reduce the amount of demonstrations a human must provide.

One major difficulty in training end-to-end robotic systems is the scarcity of data. Because human effort is often a constraint during data collection, it is desirable to collect the most useful data possible on each trial. For training purposes, high-quality data should include both successful trials \emph{and} corrective behaviors that show how to recover from bad states.
 
Since there exist some states the learning agent should never visit (such as crashes in autonomous driving tasks), it is important to explore bad states in a controlled manner. 
Given the non-i.i.d. nature of inputs to robotic control tasks, early errors often propagate throughout task execution, which leads to compounding errors and a qudratically growing regret bound in the time horizon of the task, as shown in the work of Ross and Bagnell \cite{ross2010efficient}.
By identifying the point of departure from the optimal policy, a system can target for corrective behaviors. 

Ross et al.~\cite{ross2011reduction} presented how imitation learning can be reduced to no-regret online learning by randomly switching control between the learning agent and the human demonstrator during task execution. Laskey et al. ~\cite{laskey2017learning} have recently shown the benefit of injecting control noise into optimal demonstration in order to learn corrective behaviors.
However, these methods do not leverage the input state to directly reason about whether it is potentially useful data for improving the performance of an underlying model.

In this work, we propose an active online imitation learning algorithm for deep end-to-end control systems. Our method utilizes predictive uncertainty to anticipate mistakes and switches control to the human expert at an anticipated mistake state to prevent visiting a series of bad states. Given an initial model, our proposed method will allow an imitation learning agent to minimize the number of sub-optimal states visited while still collecting labeled data at potentially interesting states. Without making unnecessary mistakes, the imitation learning agent is then able to collect more useful data given the same amount of demonstration time. As an on-policy learning algorithm as the method proposed by Ross et al.~\cite{ross2011reduction}, our method also shares the same no-regret guarantees with online data aggregation algorithms of the same kind.
Our experiments demonstrate with an end-to-end autonomous driving system that, given the same amount of data collection time and human effort, our proposed system improved performance of an imitation learning model more than the alternative methods.

\section{RELATED WORK}

Our work builds on recent advances in predictive uncertainty estimation for deep networks and is closely related to the field of imitation learning.

\subsection{Uncertainty Estimation for Deep Networks}

In machine learning, uncertainty of a point prediction has two major sources: the inherent data distribution\footnote{In certain literature~\cite{osband2016risk,osband2016deep}, the stochasticity from data distribution is referred to as \textit{risk} instead of uncertainty.} and the model parameters themselves. Uncertainty inherent in the data distribution, or \textit{aleatoric} uncertainty~\cite{kendall2015bayesian,gal2015bayesian}, will not be explained away with more data but can be explicitly modeled with various techniques~\cite{bishop1994mixture,tang2013learning}. Uncertainty in the model parameters, or \textit{epistemic} uncertainty~\cite{kendall2015bayesian,gal2015bayesian}, is reducible given infinite training data and can be used to detect adversarial inputs\footnote{The definition of adversarial inputs here refers to any input located outside the support of training data, which is slightly different from the definition in the area of adversarial training~\cite{ganin2016domain,shrivastava2017learning,miyato2016adversarial}}, i.e. where the model needs more data. We refer readers to the work of Kendall et al.~\cite{kendall2017uncertainties} for a deeper background on predictive uncertainty in deep neural networks. 

Monte Carlo (MC)-dropout~\cite{gal2015bayesian,gal2017deep} and ensembles~\cite{dietterich2000ensemble,lakshminarayanan2017simple} are two popular methods that can be used to estimate \textit{epistemic} uncertainty in deep networks. 
Dropout~\cite{srivastava2014dropout} and ensembles~\cite{dietterich2000ensemble} were discovered as regularization techniques to improve generalization performance of deep networks. The two methods share similarity in the sense that both apply probabilistic reasoning on the network weights and dropout can be interpreted as an averaged combination of ensemble models.
Recent work~\cite{gal2015bayesian,gal2017deep} has found that training with dropout is approximately performing Bernoulli variational inferences on the network weights, and therefore applying dropout at test time approximates Monte Carlo sampling from the posterior distribution of the network weights.

Our experiments in this work employ MC-Dropout for sampling output of a regression network. However, our uncertainty estimation technique for end-to-end control systems also works with ensemble outputs and our data aggregation algorithm can be easily modified to work with any uncertainty estimation mechanism.

\subsection{Data Aggregation for Imitation Learning}

In the context of imitation learning, DAgger \cite{ross2011reduction} is a popular no-regret framework for aggregating training data under the learned policy's state distribution by switching control between the learning agent and the expert. Several recent work has explored how to incorporate error prediction and uncertainty estimation into the DAgger framework. 
Zhang and Cho presented SafeDAgger \cite{zhang2016query}, adopting a safety policy for predicting errors and selecting only a subset of samples to query for labels efficiently. However, the safety policy requires a separate set of training data, which may not be readily available for real-world applications. Closely related to our work, Lee et al. \cite{lee2018safe} recently proposed a DAgger-based learning algorithm that leverages network uncertainty to effectively imitate a model predictive control (MPC) policy. By contrast, instead of learning from a MPC policy that is self-consistent and can be queried without safety concerns, our proposed system is designed to learn from human experts.

Laskey et al.\cite{laskey2017learning} proposed an off-policy imitation learning algorithm DART, which injects noise to the expert's control during data collection to approximate the learning agent's error. The resulting policy achieves better performance than that of DAgger without forcing the expert to visit a lot of sub-optimal states.
However, both DAgger and DART do not actively reason about the learned model's confidence given an input. Our proposed algorithm instead leverages uncertainty estimations during on-policy data collection to allow the learning agent to switch control at its uncertain states and thus focusing on collecting data targeted at corrective behaviors at the boundary of optimal and sub-optimal states. The comparison between our proposed method and existing data aggregation algorithms is depicted in Figure~\ref{fig:comparison}. 

Our work is also related to systems that can identify when they do not know a correct response, characterized as the Knows-What-it-Knows (KWIK) framework \cite{li2008knows}. The most closely related of these works is Confidence-Based Autonomy (CBA) \cite{chernova2009interactive}, which is an interactive imitation learning algorithm that reasons about the agent's confidence on an input to decide whether it should request a demonstration or not. CBA explicitly measures distances between data points and utilizes classification confidence as well as decision boundaries for determining whether labeling an input will be useful for the learning agent. Instead of on-policy learning, CBA requests label on demand, which may not be suitable for high-frequency decision making tasks with continuous state-action spaces such as driving.

\section{METHODOLOGY}

Given an end-to-end continuous control task and an initial model, it is desirable to improve the model's performance with as few data points as possible while keeping the expert from visiting a series of sub-optimal states. Our proposed on-line data aggregation algorithm has two major components: uncertainty estimation and active data acquisition. 

\subsection{Uncertainty Estimation}
Given an input (image and/or measurements) $\mathbf{x}$, an end-to-end control system outputs normalized continuous signals $\mathbf{y}$ (e.g. steering angle of a car).  Applying MC-Dropout, a set of output samples can be drawn for the same input $\mathbf{x}$ from multiple forward passes, with which we can obtain a discrete distribution \footnote{The granularity of discretization is problem-specific and should be balanced with the number of samples drawn.} of the output samples. A desirable uncertainty score should capture the level of inconsistency in this discrete sample distribution from all aspects.

Let $\{\mathbf{y}^n\}$ denote the set of discretized output samples drawn from $n$ forward passes, $c$ denote a class and  $ c^* $ denote the mode (which can be used as the actual control output).  Entropy $\textbf{H}$ and variational ratio $\textbf{VR}$ are two important measures for capturing categorical uncertainty \cite{gal2017active}: 

\begin{equation} \textbf{H}(\{\mathbf{y}^n\}) = -\sum_c \frac{\sum_n \mathbf{1}[\mathbf{y}^n,c]}{N}  \text{log} [\frac{\sum_n \mathbf{1}[\mathbf{y}^n,c]}{N} ]\end{equation}
\begin{equation} \textbf{VR}(\{\mathbf{y}^n\},c^*) =1 - \frac{\sum_n \mathbf{1}[\mathbf{y}^n,c^*]}{N} 
\end{equation} 
\,\,\,\, where $\mathbf{1}$ denotes the indicator function.

At the same time, end-to-end control models operate on time-series input and output continuous control signals. Therefore, we also compute the standard deviation (from the mode) $\textbf{SD}$ and temporal divergence $\textbf{TD}$ of $\{\mathbf{y}^n\}$, i.e. the KL-divergence between the output distribution of time step $k$ and $(k-1)$. Let $\{\mathbf{y}^n\}_k$ be the output set at time step $k$ (conditioned on input $\mathbf{x}_k$):

 \begin{equation} \textbf{SD}(\{\mathbf{y}^n\},c^*) = \frac{\sum_n||\mathbf{y}^n-c^*||_2}{N} \end{equation} 
\begin{multline} \textbf{TD}({\{\mathbf{y}^n\}}_k,{\{\mathbf{y}^n\}}_{k-1}) = 
\textbf{KL}[p({\{\mathbf{y}^n\}}_k)||p({\{\mathbf{y}^n\}}_{k-1})] \end{multline}

 Combining the above measures, our proposed uncertainty score describes the level of inconsistency in the output sample distribution by taking into account categorical uncertainty, temporal smoothness and the expected error in the output distribution:
\begin{multline} 
\textbf{U}({\{\mathbf{y}^n\}}_k,{\{\mathbf{y}^n\}}_{k-1}) = \\ [\textbf{TD}({\{\mathbf{y}^n\}}_k,\mathbf{\{\mathbf{y}^n\}}_{k-1})  \cdot 
  \textbf{H}({\{\mathbf{y}^n\}}_k) \cdot \textbf{VR}({\{\mathbf{y}^n\}}_k,c^*) \\
  + \lambda  \textbf{SD}({\{\mathbf{y}^n\}}_k,c^*) ]^2
\end{multline}

Empirically \footnote{Note that the exact form of the uncertainty score function can be domain-specific and network-specific.}, $\textbf{TD}$, $\textbf{VR}$ and $\textbf{H}$ values are noisy when used alone and therefore are multiplied as one term in the uncertainty score function. The $\lambda$ term is used to weigh $\textbf{SD}$ such that all the terms are on the same order of magnitude. 
Applying a quadratic filter helps to further reduce noise and balance false-positive and true-positive rates.

\begin{algorithm}[t!]

{\caption{UAIL (\textbf{Input}: Environment $P$, Initial Demonstrations $D_0$, Expert Policy $\pi^*$, Uncertainty Threshold $\eta$, Time Window $T$, Sample Size $N$, Learning Episodes $E$, Batch Size $B$; \textbf{Output}:  Policy $\pi$)} \label{alg:UAIL}}
{
\begin{itemize}
  \item Initialize demonstration set $D = D_0$;
  \item \textbf{Repeat} for $E$ times:
  \begin{enumerate}
  \item Train neural network policy $\pi$ with $D$;
  \item Sample initial state $s_0$ from $P$ and set $t=0$;
  \item Initialize Uncertainty Array $U[T]$;
  \item \textbf{while} size of $D$ is less than  $B$:
  \begin{enumerate}
  \item Obtain ${\{\mathbf{y}^n\}}_t$ with MC-Dropout on $s_t$;
  \item Compute $U[t \bmod T]=\textbf{U}({\{\mathbf{y}^n\}}_t, {\{\mathbf{y}^n\}}_{t-1})$ ;
  \item $D = D\cup \{s_t, \pi^*(s_t)\}$;
  \item \textbf{if} $\sum_{k=0}^T U[k] > \eta$: $s_t = P(s_t, \pi^*(s_t))$;
  \item \textbf{else}: $s_t = P(s_t, \pi(s_t))$;
  \end{enumerate} 
  \end{enumerate}
  \item return $\pi$
\end{itemize}
}
\end{algorithm}

\subsection{Active Data Acquisition}

In imitation learning, experts often demonstrate only optimal actions and therefore seldom visit sub-optimal states. However, with limited training data and non-i.i.d. inputs, the learning agent is bound to make mistakes and visit adversarial states that are sub-optimal. It is desirable to also collect action labels at these adversarial states.

DAgger \cite{ross2011reduction} addresses this issue by switching controls in between the learning agent and the human expert at random during task execution and collecting only the human's control signals. However, random control switches often makes it hard for humans to demonstrate naturally due to the sparsity of actual feedback. Laskey et al. proposed DART \cite{laskey2017learning}, which, instead of forcing the demonstrator to visit sub-optimal states under the agent's policy, approximates the noise in the learned policy during off-policy imitation learning. DART utilizes control noise to explore the boundary between \textit{good} and \textit{bad} states during data collection. However,  collection process can be made more effective by actively detecting adversarial states.

With uncertainty estimations, a learning agent can now predict when it is likely to make a mistake and switch control to the human expert in order to prevent visiting a series of sub-optimal states. We propose an uncertainty-based data aggregation algorithm named UAIL (Uncertainty-Aware Imitation Learning), which detects adversarial states actively in order to fix the learning agent's mistake as soon as possible. 
As shown in Algorithm~\ref{alg:UAIL}, per-frame uncertainty in a short time window $T$ is accumulated for estimating the total uncertainty at time $t$ to decide whether the agent should switch control to the human expert. The action that the expert takes is recorded as the optimal action for all input frames. The data collection and model training process alternates.

\section{EXPERIMENT SETUP}

We tested our method in end-to-end autonomous driving domain. Existing end-to-end driving networks have shown their ability to perform road-following and obstacle avoidance~\cite{pomerleau1989alvinn,muller2006off,bojarski2016end,bojarski2017explaining}. Recent work has been exploring how to leverage these capabilities for practical use. Codevilla et al.~\cite{codevilla2017end} proposed to use a  command-conditional network to address the ambiguity~\cite{pomerleau1989alvinn} in learning the optimal action to take at intersections. The learned model can then be combined with a high-level planner that issues route commands. We employed this model for evaluating our uncertainty estimation technique and data aggregation algorithm on autonomous driving tasks. We conducted experiments in the CARLA 3D driving simulation environment \cite{dosovitskiy17}.

\subsection{Uncertainty Estimation}
While the performance of MC-Dropout for uncertainty estimation has been evaluated in prior work~\cite{gal2015bayesian,gal2017active,lee2018safe}, leveraging such uncertainty estimation for predicting infractions in temporal decision making tasks, to the best of our knowledge, has not been previously explored.  

Our proposed uncertainty estimation technique was evaluated on an existing autonomous driving model provided by Codevilla et al.~\cite{codevilla2017end}. The model was trained on two hours of human-driving data collected in simulation in \textit{Town 1} in the CARLA environment. The imitation network takes image of the front camera and the speed of the vehicle as input and outputs steering angle and throttle value. 
The imitation agent and our uncertainty estimation system are tested in a novel environment (\textit{Town 2}) with both seen and unseen weather conditions using a subset\footnote{Our test set focuses on cases where the provided model performs poorly, i.e. has one or more infractions across trials. } of test cases provided in the CARLA benchmark \cite{dosovitskiy17}. Collisions, intersections with the opposite lane, and driving onto the curb are recorded as infractions. 
The network outputs two control signals: steering angle and throttle value. Uncertainties for the two control signals were computed independently and summed with weights in the total uncertainty estimation function. The tested uncertainty estimation signals are: \par
1) \textit{Steer Error} $\textbf{SD}_{steer}$; \par
2) \textit{Throttle Error} $ \textbf{SD}_{throttle}$; \par
3) \textit{Total Uncertainty} $(\textbf{U}_{steer} + \alpha \textbf{U}_{throttle})$; \par

In our tests, 20 output samples were used per input. The value of $\alpha$ was set empirically as $0.6$.

\subsection{Active Data Acquisition}

For an end-to-end control task, we hypothesize that: 
\begin{enumerate}
\item Given a pool of training data, the subset selected with our uncertainty estimations will improve an agent's performance more than a subset selected randomly; 
\item Given same amount of data collection time and human effort, data collected with UAIL will improve an agent's performance more than data collected with alternative methods.
\end{enumerate}
To test these hypotheses, we obtained the set of demonstration data provided by Codevilla et al. \cite{codevilla2017end} and cleaned it up by removing data files that contained infractions. We refer to this data set as the \textit{passive} dataset. We selected a subset of data files from the clean data set and used it as the \textit{starter set} from which we will improve the trained model's performance using different data selection methods. 

To test hypothesis 1), we randomly sampled a fixed amount of data from \textit{passive} and added it to the \textit{starter set} to serve as a \textit{baseline} for further comparison. 
We then processed the \textit{passive} dataset using our proposed uncertainty scoring function and obtained a set of data named \textit{active filter} by sorting all the data files by their accumulated uncertainty and adding the top ones to \textit{starter set} such that \textit{active filter} has the same size as that of \textit{baseline}. 

To test hypothesis 2), we collected new demonstration data in \textit{Town 1} by recording human drivers operating the simulated car (using a Logitech G29 steering wheel controller) under three different conditions: \textit{stochastic mixing}, \textit{random noise}, and \textit{UAIL}, where \textit{stochastic mixing} and \textit{random noise} are the one-step versions of DAGGER and DART. A total number of 12 participants\footnote{11 out of 12 participants have a US-issued driving license and 1 has a learner's permit.} contributed to the driving data, which amounts to 2 hours of driving per condition. The obtained three different datasets are all of the same size as \textit{baseline} and their compositions are:
\begin{itemize}
\item \textit{Stochastic mixing}: \textit{Starter Set} and newly collected data with 40\% agent control at random;
\item \textit{Random noise}: \textit{Starter Set} and newly collected data with injected random noise within \ang{30} of the ego-vehicle's current heading at every 5 frames;
\item \textit{UAIL}: \textit{Starter Set} and newly collected data with active control switch at high uncertainty states.
\end{itemize}
 
The parameters for designing the three different conditions were chosen to control the level of human effort required and were set empirically such that the agent takes control or injects noise as frequently as possible but at a level such that the vehicle is still controllable for experienced human drivers. A preliminary user study was conducted to serve as a measure for how well the level of human effort is controlled. The user study collected subjective views of the participants on how they would rank the easiness of control under the three different conditions (without knowing which is which).

We created our own \textit{Intersections} benchmark with the default maps in CARLA to extensively test the learned models' performance on handling intersections. The benchmark was designed to have a balanced number of test cases among left turns, right turns, and go-straight scenarios at intersections.

\begin{table}[b]
\vspace{-0.4cm}
\centering
\renewcommand{\arraystretch}{1.4}
\begin{tabular}{c c c c c c c }
\hline
 \multicolumn{3}{c}{\textbf{Conditions}} & \multicolumn{4}{c}{\textbf{Median Uncertainty Value}}\\
 Map & Weather & Agents & Follow & Left & Right & Straight \\
\hline
\hline
Town1 & Seen & No & 0.694 & 0.827 &  0.868 & 0.667 \\
Town1 & Seen & Yes & 0.739 & 0.823 & 0.854 & 0.717 \\
Town1 & Unseen & Yes & 0.737 & 0.870 & 0.903 & 0.757 \\
Town2 & Seen & Yes & 0.759 & 0.852 &0.881 & 0.815 \\
Town2 & Unseen & Yes & 0.740 & 0.809 & 0.952 & 0.753 \\
\hline
 \multicolumn{3}{c}{ Map Town1 Avg.} & \multicolumn{4}{c}{0.788} \\
 \multicolumn{3}{c}{ Map Town2 Avg.} & \multicolumn{4}{c}{0.820} \\
 \multicolumn{3}{c}{ Seen Weather Avg.} & \multicolumn{4}{c}{0.791} \\
 \multicolumn{3}{c}{ Unseen Weather Avg.} & \multicolumn{4}{c}{0.815} \\
\hline
\end{tabular}
\caption{\small Median Uncertainty Value in Different Scenarios}
\label{evaluation}
\vspace{-0.4cm}
\end{table}

\begin{figure*}[ht]
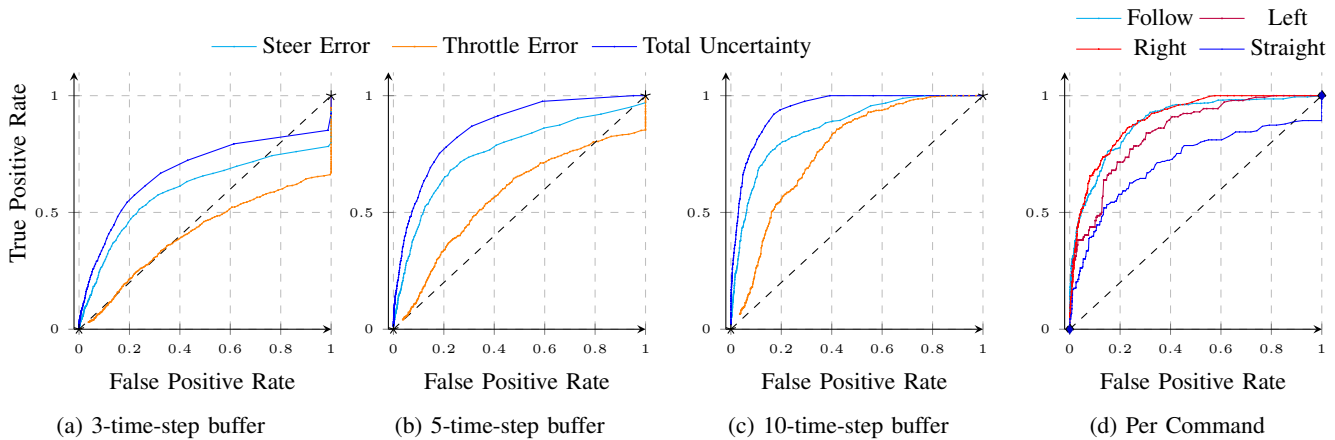

\input{x_3steps.tex}
\input{x_5steps.tex}
\input{x_10steps.tex}
\input{x_roc_by_command.tex}
\vspace{-0.5cm}

\caption{ROC curves for predicting infractions in test environment:(a)(b)(c) plot different uncertainty functions under different time-step buffers; (d) plots the total uncertainty function under different commands with 5 time-step buffer.}
\label{fig:roc_curves}
\vspace{-0.4cm}
\end{figure*}

\section{RESULTS}

\subsection{Uncertainty Estimation}

We evaluated how well our proposed candidate uncertainty functions predict an infraction by plotting their receiver operating characteristic (ROC) curves. Since delays are expected in between a high-uncertainty estimation and an infraction, we employed time-step buffers to account for the variable delay when evaluating the predictions. A $k$-time-step buffer will allow any prediction found to be less than $k$ time-steps ahead an infraction to be counted as a true positive. Time-step buffers with 3, 5, and 10 time-steps\footnote{In our simulated experiments, with 20 MC-Dropout samples, the FPS is around 3 and therefore 3 time-steps are about 1 second. (The low FPS was due to running both the driving network and CARLA simulation on a local machine.)} are used to evaluate different uncertainty functions. Since the network has a branched structure, ROC curves under different commands are plotted in Figure~\ref{fig:roc_curves}(d). The ROC curves under different commands have different shapes, which indicates that different threshold values should be used to achieve similar true-positive ratios across different commands.

\begin{figure}[h]
\centering
\begin{subfigure}[b]{0.98\linewidth}
\centering
\includegraphics[scale=0.2]{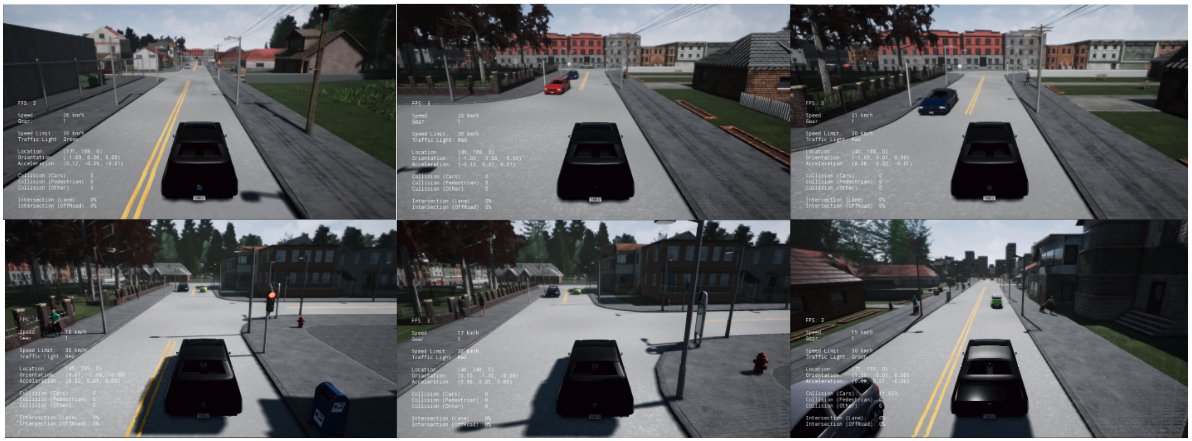}
\caption{\small Selected frames with estimated uncertainty below threshold tend to correspond with scenarios in which the agent has collected many training data.}
\label{fig:lu}
\end{subfigure}
~
\begin{subfigure}[b]{0.98\linewidth}
\vspace{0.2cm}
\centering
\includegraphics[scale=0.2]{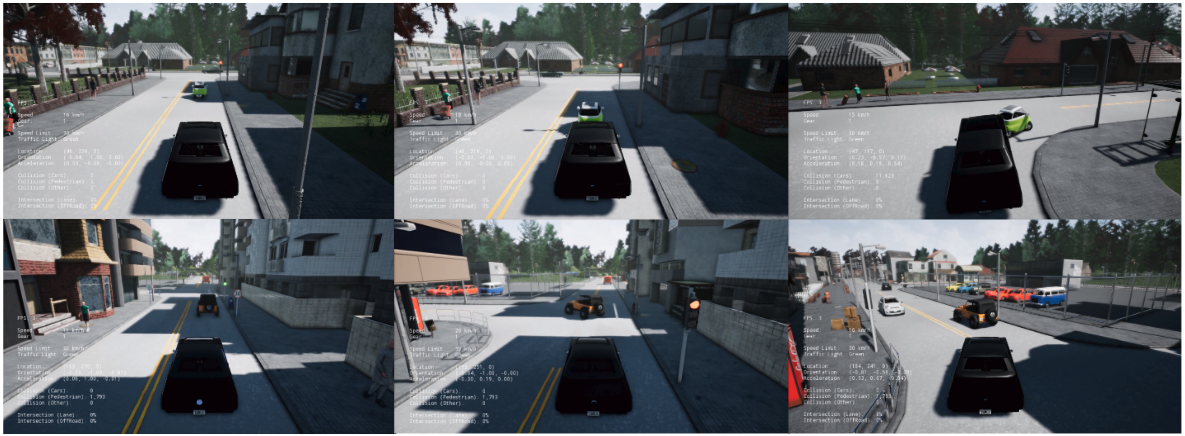}
\caption{\small Selected frames with estimated uncertainty above threshold tend to correspond with scenarios containing lighting changes, unseen agents or infractions.}
\label{fig:hu}
\end{subfigure}
\caption{\small Example CARLA frames from on-line uncertainty monitoring. (Note: these are not input to the network but CARLA graphical displays. The network takes input from a front facing camera mounted on the car.)}
\label{fig:monitor}
\vspace{-0.3cm}
\end{figure}

\begin{figure}
\begin{subfigure}[b]{0.92\linewidth}
\centering
\includegraphics[scale=0.25]{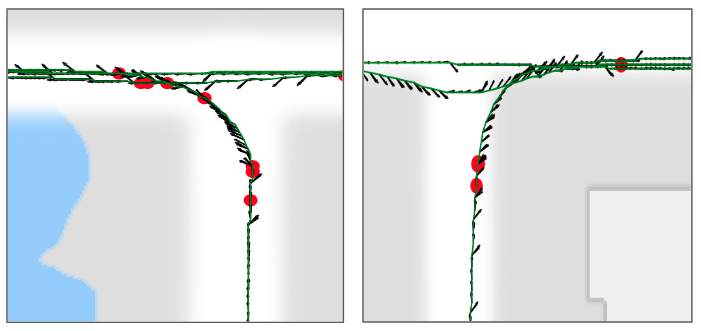}
\caption{Confident Turns}
\label{fig:ct}
\end{subfigure}
~
\begin{subfigure}[b]{0.9\linewidth}
\vspace{0.2cm}
\centering
\includegraphics[scale=0.25]{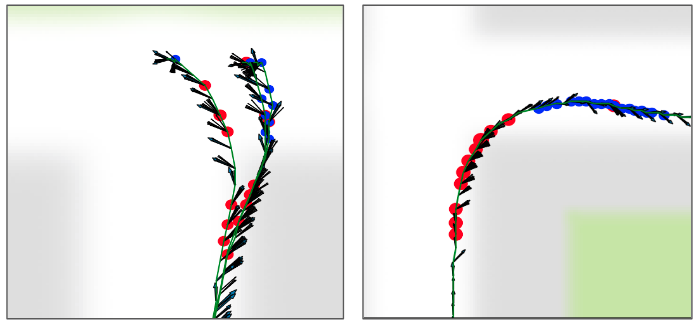}
\caption{Uncertain Turns}
\label{fig:ut}
\end{subfigure}
\caption{\small Map view of annotated trajectories: \textit{green} lines are the agent's trajectories; \textit{red} circles indicate where uncertainty exceeded threshold; \textit{blue} circles indicate infractions; small \textit{black} arrows denote the MC-dropout samples of steering angles. }
\label{fig:monitor2}
\vspace{-0.2cm}
\end{figure}

 Figure~\ref{fig:roc_curves} shows the ROC curves for the candidate signals. The smaller the time-step buffer is, the more likely an estimation will be treated as false positive, in which cases leveraging past estimations could help predicting infractions. As indicated by the area under the ROC curves, our proposed uncertainty estimation function outperforms raw standard deviation measures (i.e. Steer and Throttle errors).

Given a desired true-positive/false-positive ratio, an uncertainty threshold can be selected for online monitoring purpose. Figure~\ref{fig:monitor} shows example frames that are below or above the selected threshold during online monitoring. 
Figure~\ref{fig:monitor2} shows 2D-map projections of example trajectories of the ego-vehicle during turning behaviors with annotated locations at which online uncertainty measure surpassed threshold or actual infractions happened. 

To test if the uncertainty estimation is sensitive to novel scenes, we collected five datasets under different scenarios (i.e. training/testing map, seen/unseen weather etc.), each consisting of 9,600 frames. The median uncertainty measures under different commands are shown in Table~\ref{evaluation}. The general trend is that frames taken in novel environment and under unseen weather have a higher average uncertainty value than those from seen weather and environment. The dataset containing no other agents (cars or pedestrians) has the lowest uncertainty value under most commands.

\subsection{Active Learning}

\begin{figure}[t]
\centering
\includegraphics[scale=0.43]{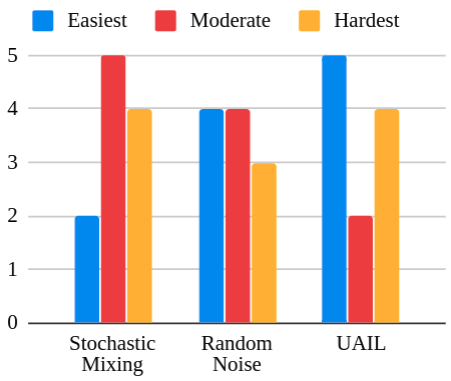}
\caption{\small Histogram of responses to user study question: \textit{How would you rank the three conditions by easiness of driving?} (p-values obtained from performing t-test for the three conditions are 0.481, 0.741 and 0.770 respectively.)}
\label{fig:user_study}

\vspace{-0.4cm}
\end{figure}

\begin{table}[t] 
\centering
\renewcommand{\arraystretch}{1.4}

\begin{tabular}{l c c c c c}
\hline
\multirow{2}{*}{\textbf{Dataset}} &  \multirow{2}{*}{\textbf{Infraction}} &\multicolumn{2}{c}{\textbf{Success Rate}} & \multicolumn{2}{c}{\textbf{Km per Infraction}}\\
&  \textbf{Rate} & Town1 & Town2 & Town1 & Town2 \\
\hline
\hline
\multicolumn{1}{l}{Passive (full)} &  -   & 0.55 & 0.40 &  0.69 & 0.55 \\
\multicolumn{1}{l}{Baseline}  & -  & 0.52 & 0.34 &  0.90 & 0.47 \\
\multicolumn{1}{l}{Starter Set} & -  & 0.41 & 0.24 & 0.65 & 0.56 \\
\multicolumn{1}{l}{Active Filter} & -  & 0.68 & 0.51 &  0.90 & 0.67 \\
\hline
\multicolumn{1}{l}{Stochastic Mix} &  20.05 \% & 0.58 & 0.44 & 0.83 & 0.47 \\
\multicolumn{1}{l}{Random Noise}  &  19.54 \% &  0.73 & 0.51 & 0.75 & 0.51 \\
\multicolumn{1}{l}{UAIL} & \textbf{ 13.83 \% } & \textbf{0.74} &  \textbf{0.61} & 0.88  &  0.63 \\
\hline
\end{tabular}
\caption{\small Performance Comparison on \textit{Intersections} Benchmark. Avg success rate and distance traveled between infractions are reported. Distance traveled between infractions can be higher for a model with lower success rate due to its failure in learning turning behaviors.}
\label{performance}
\vspace{-0.2cm}
\end{table}

The performance\footnote{Video demonstrating uncertainty monitoring and behavior of the trained agents can be found at \url{https://youtu.be/I6z176kr1ws}} of the models trained with different datasets is shown in Table~\ref{performance}. As expected, all models improved after incorporating additional data. The model trained with the \textit{active filter} dataset outperformed that with \textit{baseline} as we hypothesized. Interestingly, the model trained with \textit{active filter} also achieved better performance than the model using all the \textit{passive} data, which suggests that in certain cases, likely when the training data distribution has multiple modes (different driving styles in this case), less data can train an agent with better behavior. 

Among the three newly collected datasets, the model trained with UAIL data has the highest success rates, and at the same time the infraction rate of the data collected with UAIL is the lowest, which indicates that it is safer to collect data with UAIL than using alternative methods. As shown in Figure~\ref{fig:user_study}, our user study did not indicate any method to be significantly more difficult than the others across users. We believe we were able to control the level of human effort required at an even level and therefore evaluating the models trained with data obtained under the selected three different conditions is a fair comparison for the algorithms under test.
Therefore, our primary experimental results agree with our hypothesis, demonstrating that UAIL can be used to collect more useful data for improving the performance of an initial model given the same amount of data collection time and human effort.

\section{CONCLUSION}

In this paper, we present a technique to estimate uncertainty for end-to-end control systems and show how such estimation can be leveraged to predict infractions and acquire new training data selectively. 
We demonstrate in an end-to-end autonomous driving system, that our proposed system allows an imitation learning agent to selectively acquire new input data from human experts at states with high uncertainty in order to maximally improve its performance.

One limitation of data aggregation methods like ours is that they require new expert demonstrations, which may not be available/preferable in certain use cases. Future extension of this work may include examining how to leverage uncertainty estimations as self-supervision signals and improve the agent's learned policy through reinforcement learning.
 
 In our pilot user study, we observed a bi-modal distribution for user's experience with UAIL, which will require an in-depth investigation by explicitly measuring participant's locus of control \cite{lefcourt1991locus} and taking participant's past experiences (e.g. proficiency in driving and/or playing video games) into consideration in order to to conclude whether UAIL is more efficient for a certain type of users.




\section*{ACKNOWLEDGMENT}
This work is a collaborative effort of Honda Research Institute, US (HRI-US) and the Personal Autonomous Robotics Lab (PeARL) at The University of Texas at Austin. PeARL research is supported in part by the NSF (IIS-1724157, IIS-1638107, IIS-1617639, IIS-1749204) and ONR(N00014-18-2243).



\bibliography{1_root.bbl} 
\bibliographystyle{plain}

\end{document}